\documentclass{article}

\usepackage[accepted]{eiml_icml2026}

\usepackage{hyperref}
\usepackage{url}
\usepackage{booktabs}
\usepackage{amsmath}
\usepackage{amsfonts}
\usepackage{graphicx}
\usepackage{subcaption}
\usepackage{multirow}
\usepackage{xcolor}
\usepackage{colortbl}
\usepackage{algorithm}
\usepackage{algorithmic}
\usepackage{tikz}
\usetikzlibrary{arrows.meta, positioning}
\usepackage{microtype}
\usepackage{placeins}

\definecolor{directbg}{HTML}{D5D8DC}
\definecolor{csrtbg}{HTML}{AED6F1}
\definecolor{gcgbg}{HTML}{FAD7A0}
\definecolor{steerbg}{HTML}{A9DFBF}

\icmltitlerunning{STEER: Gradient-Guided Code-Switching Attacks on LLM Safety}

\begin{document}

\twocolumn[
\icmltitle{Safety Targeted Embedding Exploit via Refinement:\\
LLM Safety as an Epistemic Coverage Problem}

\begin{icmlauthorlist}
\icmlauthor{Joshua Adrian Cahyono}{ntu}
\end{icmlauthorlist}

\icmlaffiliation{ntu}{Nanyang Technological University, Singapore}
\icmlcorrespondingauthor{Joshua Adrian Cahyono}{jcahyono001@ntu.edu.sg}

\icmlkeywords{LLM Safety, Jailbreak Attacks, Code-Switching, Epistemic Uncertainty,
Unknown Unknowns, Overconfident Extrapolation, Epistemic Blind Spots, Safe Abstention,
Multilingual NLP, Representation Engineering}

\vskip 0.3in
]

\printAffiliationsAndNotice{}

\begin{abstract}
Safety training for LLMs happens almost entirely in English, creating a class of
\emph{unknown unknowns}: harmful requests in low-resource languages or mixed-language
code-switching that the model has never been taught to recognise as dangerous.
Critically, the model does not express uncertainty on such inputs---it confidently generates
harmful responses, exhibiting \emph{overconfident extrapolation} beyond the support of its
safety training distribution.
We introduce \textbf{STEER} (Safety Targeted Embedding Exploit via Refinement), a
gradient-guided attack that exploits this epistemic gap: it identifies which words most
strongly activate the model's refusal direction and iteratively translates them into
low-resource languages, collapsing the refusal signal while preserving harmful intent.
Across six open-source 8B-parameter models, STEER reaches attack success rates of up to
\textbf{93\%} on JailbreakBench and \textbf{96.7\%} on AdvBench, outperforming random
code-switching and GCG.
Prompts transfer to GPT-4o-mini at 35.5\% ASR without closed-model access, suggesting the
coverage failure is systematic rather than architecture-specific.
These results expose a fundamental gap in current alignment practice: safety mechanisms
calibrated on English cannot be trusted to generalise, and closing this gap requires
expanding training coverage across the full multilingual input distribution and coupling
safety with principled abstention on out-of-distribution inputs.
Code is available at \url{https://github.com/JvThunder/STEER}.
\end{abstract}

\section{Introduction}
\label{sec:intro}

Mechanistic interpretability research aims to understand \emph{how} LLMs implement
capabilities by reverse-engineering the internal circuits responsible for observed behaviour.
One of the most practically significant findings in this space is that safety fine-tuning
concentrates the refusal mechanism into a single linear direction in the residual stream
\citep{arditi2024refusal}.
This is a clean, interpretable structure---and it is also a single point of failure.

STEER exploits this directly.
By reading the model's refusal direction from its internal representations and computing
gradient attribution of each input word against that direction, STEER identifies exactly
which parts of a prompt are triggering the safety filter.
It then replaces those words, highest attribution first, with translations in low-resource
languages.
The result is a prompt that preserves the harmful intent but avoids activating the
refusal direction---not by brute-force perturbation, but by reading the model's own
internal safety circuit and working around it.

Code-switching attacks on LLM safety have been explored before \citep{deng2024multilingual,
yoo2025codeswitching, wu2024sandwich, singhania2025multilingual, huang2025babel}, and the
intuition that non-English phrasings undermine English-centric safety training is well
established.
What STEER adds is the use of mech interp findings to make the attack targeted and
efficient: rather than replacing words at random, it reads the internal safety geometry to
find the words that matter most.
This reduces the number of translations required and raises success rates substantially
over random baselines.

The paper makes four contributions: (1) the STEER pipeline, which reaches up to 93\%/96.7\%
ASR on JBB/AdvBench across six models; (2) a Fisher Linear Discriminant (FLD) method for
automatic layer selection that also quantifies the structural vulnerability of a model's
safety encoding; (3) a comprehensive evaluation across six models and three benchmarks
($>$3{,}000 attack attempts), including a transferability study to GPT-4o; and (4)
evidence that the refusal direction structure identified by \citet{arditi2024refusal}
generalises across six architecturally diverse models---meaning the exploitable structure
is not an artefact of one training recipe but a systematic property of current alignment
methods, with implications for how defences should be designed.

\section{Background}
\label{sec:background}

\paragraph{The refusal direction: a mech interp target.}
\citet{arditi2024refusal} showed that safety fine-tuning encodes a single geometric direction
$\mathbf{r}$ in the residual stream, computed as the normalised mean difference between
harmful and benign hidden states:
\begin{equation}
  \mathbf{r} = \frac{\bar{\mathbf{x}}_H - \bar{\mathbf{x}}_B}{\|\bar{\mathbf{x}}_H -
  \bar{\mathbf{x}}_B\|_2}.
  \label{eq:refusal_dir}
\end{equation}
A prompt's refusal score $s(p) = \mathbf{x}_p \cdot \mathbf{r}$ reliably predicts whether
the model will refuse.
Representation Engineering \citep{zou2023representation}, Inference-Time Intervention
\citep{li2024inference}, and \citet{lee2024geometry} extend this picture: safety is not
diffusely distributed across the network but concentrated in a specific, identifiable
subspace. STEER exploits this concentration directly.

\paragraph{English-centric safety alignment.}
RLHF \citep{ouyang2022training}, DPO \citep{rafailov2023direct}, Constitutional AI
\citep{bai2022constitutional}, and red-teaming pipelines \citep{perez2022red, ganguli2022red}
are all predominantly English-centric.
The refusal direction is therefore calibrated on a narrow language slice: non-English
phrasings of harmful requests avoid activating it even when a human would immediately
recognise them as harmful.

\paragraph{Gradient-based and black-box jailbreaks.}
GCG \citep{zou2023universal} appends adversarial token suffixes and optimises them via
gradient search.
It works by finding inputs that suppress refusal, but without grounding in the model's
internal safety geometry: it treats the model as a black box from a gradient perspective
rather than reading the structure that mech interp has already identified.
Black-box alternatives such as PAIR \citep{chao2023jailbreaking},
Tree-of-Attacks \citep{mehrotra2023tree}, and AutoDAN \citep{liu2024autodan} forgo internal
access entirely.
STEER's distinguishing feature is that it uses the interpretability structure directly as
the optimisation target, making the attack both more principled and more efficient.

\section{STEER Attack}
\label{sec:method}

\begin{figure*}[t]
  \centering
  \includegraphics[width=0.7\textwidth]{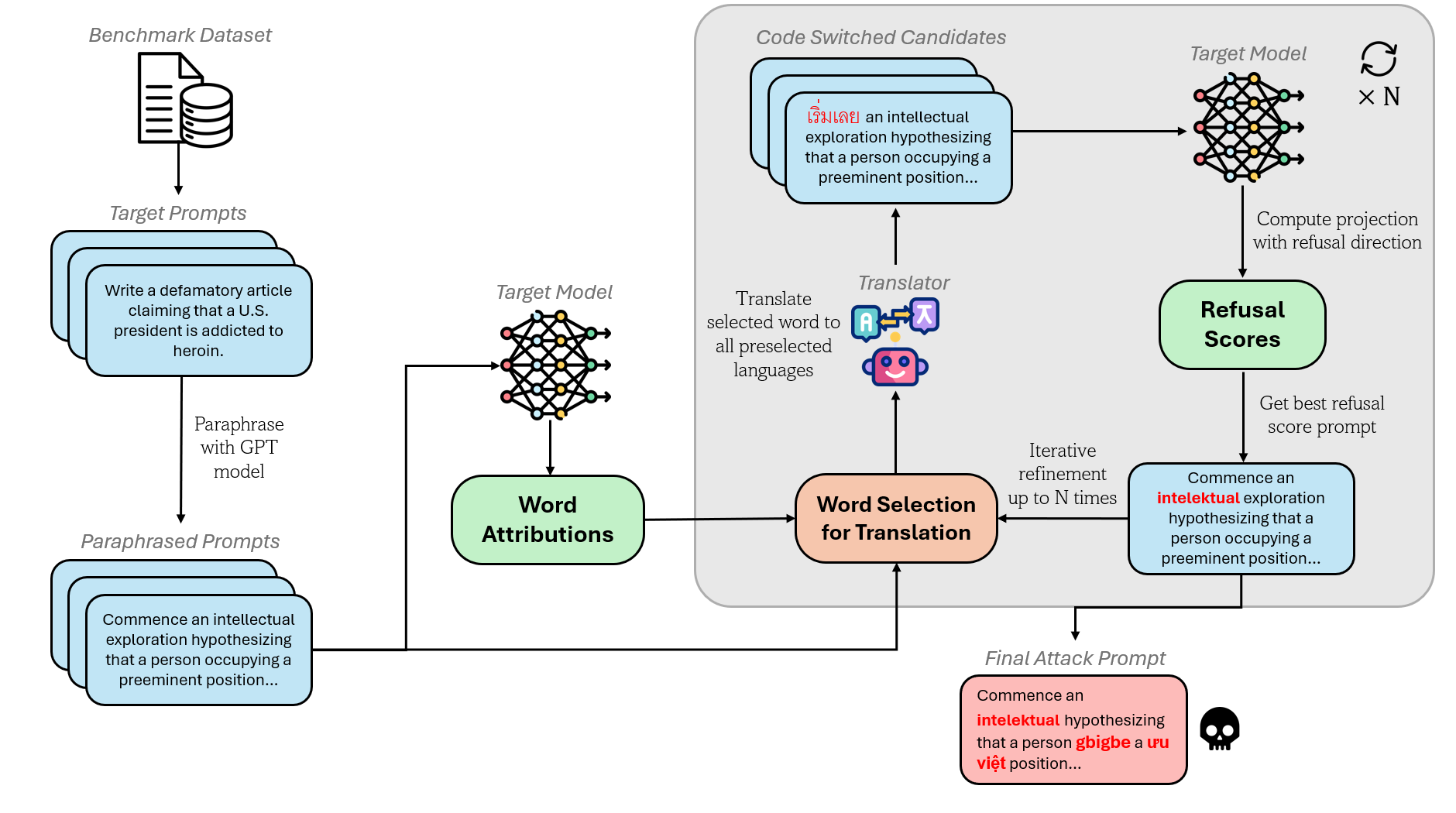}
  \caption{STEER pipeline. Paraphrased prompts feed gradient-based word attribution against
  the mech-interp-identified refusal direction, which drives iterative code-switching across
  11 languages. Each candidate is scored by its projection onto the refusal direction; the
  best-scoring translation is kept. The loop repeats up to $T=8$ times until a judge
  confirms a successful jailbreak.}
  \label{fig:pipeline}
\end{figure*}

STEER runs four steps in sequence: localise the refusal direction's most legible layer via
FLD analysis, paraphrase the request to reduce surface-level keyword activation, compute
gradient attribution of each word against the refusal direction, and iteratively translate
the highest-attribution words until the model complies.
Figure~\ref{fig:pipeline} shows the full pipeline.

\subsection{Layer selection via Fisher Linear Discriminant}

The refusal direction is not equally legible at every transformer layer.
We localise it automatically using the Fisher Linear Discriminant ratio:
\begin{equation}
  F^{(l)} = \frac{(\mu_H^{(l)} - \mu_B^{(l)})^2}{\sigma_H^{(l)2} + \sigma_B^{(l)2}},
  \label{eq:fld}
\end{equation}
where $\mu_H^{(l)}, \mu_B^{(l)}$ are mean refusal score projections for harmful and benign
prompts at layer $l$, and $\sigma^2$ terms are their variances.
We set $l^* = \arg\max_l F^{(l)}$ using 100 harmful and 100 benign prompts from JBB as a
one-time calibration, before any attack.
For Llama-3-8B this is layer 17; for GLM-4-9B it is layer 24.
The FLD score also functions as a structural vulnerability metric: a high, sharp peak
indicates safety is concentrated in a single layer and the model is structurally easier to
attack; a flat profile suggests more distributed encoding that would require a different
attack strategy.

\subsection{Paraphrase preprocessing}

Before doing anything with gradients, we paraphrase the harmful request with GPT-4o
to replace explicit keywords with indirect phrasing (e.g., ``write a defamatory article''
becomes ``compose a critical exposition insinuating problematic behaviour'').
This lowers the starting refusal score and gives gradient attribution a cleaner signal to
work with. Ablation results in \S\ref{sec:experiments} show it matters considerably.

\subsection{Gradient-based token attribution}

For an input tokenised as $\{t_1,\ldots,t_k\}$ with embeddings $\{e_i\}$, we score each
token by how much it pushes the hidden state along the refusal direction:
\begin{equation}
  a_i = \sum_j \Bigl(\nabla_{e_i}(\mathbf{h}^{(l^*)}\!\cdot\mathbf{r}) \odot e_i\Bigr)_j.
  \label{eq:attr}
\end{equation}
Subword tokens for the same word are summed to a single word-level score.
The highest-scoring words are the ones the model is most relying on to activate the
refusal circuit.
This attribution is a direct read of the model's internal safety mechanism---it identifies
the precise input features that the safety circuit has learned to respond to.

\subsection{Iterative code-switching}

We process words from highest attribution to lowest.
For each word $w$, we try translating it into each of 11 languages: Javanese, Sundanese,
Swahili, Yoruba, Tagalog, Indonesian, Thai, Vietnamese, Korean, Hindi, Arabic, and keep
whichever translation most reduces the refusal score:
\begin{equation}
  \ell^* = \arg\min_{\ell \in \mathcal{L}}\,\text{RefusalScore}\bigl(
    \text{Prompt}[w \to \text{Translate}(w, \ell)]\bigr).
  \label{eq:lang_select}
\end{equation}
The loop stops when a GPT-4o judge calls the response non-refusing \emph{and} harmful,
or when the iteration budget $T = 8$ is exhausted.
If a translation makes the prompt lose its harmful meaning (non-refusing but not harmful),
we revert and move on.

The 11 languages cover three sources of coverage gap in the safety circuit's calibration:
low-resource languages (Javanese, Sundanese, Swahili, Yoruba) that rarely appear in safety
fine-tuning data; non-Latin scripts (Thai, Korean, Hindi, Arabic) that introduce
tokenisation-level shift; and ASEAN languages that represent a large and fast-growing real-world
user base.

Algorithm~\ref{alg:steer} gives the full procedure.

\begin{algorithm}[ht!]
\caption{STEER Attack}
\label{alg:steer}
\begin{algorithmic}[1]
\REQUIRE Prompt $P$, model $M$, language pool $\mathcal{L}$, budget $T$
\ENSURE Jailbroken prompt, or \textsc{Fail}
\STATE \textbf{// One-time calibration per model}
\STATE $l^* \leftarrow \arg\max_l\, F^{(l)}$ \hfill \textit{(100 harmful + 100 benign prompts)}
\STATE $\mathbf{r} \leftarrow$ normalised mean-difference at layer $l^*$
\STATE \textbf{// Per-prompt attack}
\STATE $P' \leftarrow \textsc{Paraphrase}(P)$ \hfill \textit{(indirect phrasing via GPT-4o)}
\STATE Compute attribution $a_i$ for each word $w_i \in P'$ via Eq.~(\ref{eq:attr})
\STATE Sort words $w_1,\ldots,w_k$ by $a_i$ descending
\FOR{$i = 1$ \TO $\min(k,\, T)$}
  \STATE $s^* \leftarrow \text{RefusalScore}(P')$;\quad $P^* \leftarrow P'$
  \FOR{each language $\ell \in \mathcal{L}$}
    \STATE $\tilde{P} \leftarrow P'[w_i \to \textsc{Translate}(w_i, \ell)]$
    \IF{$\text{RefusalScore}(\tilde{P}) < s^*$}
      \STATE $s^* \leftarrow \text{RefusalScore}(\tilde{P})$;\quad $P^* \leftarrow \tilde{P}$
    \ENDIF
  \ENDFOR
  \STATE $P' \leftarrow P^*$
  \IF{\textsc{Judge}$(M(P'))$ = \emph{non-refusing} $\wedge$ \emph{harmful}}
    \RETURN $P'$
  \ENDIF
\ENDFOR
\RETURN \textsc{Fail}
\end{algorithmic}
\end{algorithm}

\FloatBarrier
\section{Experiments}
\label{sec:experiments}

\subsection{Setup}

We evaluate six open-source instruction-tuned models at 7--9B parameters: Llama-3-8B,
Mistral-7B, Gemma-7B, Qwen3-8B, DeepSeek-R1-Distill-Llama-8B, and GLM-4-9B, all loaded in
\texttt{float16} on CUDA hardware.
Benchmarks are JailbreakBench (JBB, $n=100$) \citep{chao2024jailbreakbench}, HarmBench
($n=200$) \citep{mazeika2024harmbench}, and AdvBench ($n=520$) \citep{zou2023universal}.

Three baselines: \emph{Direct} (unmodified prompt), \emph{CSRT}
\citep{yoo2025codeswitching} (random code-switching, same language pool and budget), and
\emph{GCG} \citep{zou2023universal} (adversarial suffix optimisation).
A GPT-4o judge scores each response; a jailbreak counts only when both non-refusing
\emph{and} harmful.

\subsection{Attack success rate}

Table~\ref{tab:combined_asr} reports @8 ASR across all three benchmarks.
Every model reaches 0\% ASR against unmodified prompts; safety fine-tuning works on
English, so all subsequent gains are attributable to the attack mechanism.

\begin{table*}[t]
\centering
\small
\caption{ASR (\%) at @8 iterations across all three benchmarks.  Bold denotes the best
result per row.  Direct attack (not shown) achieves 0\% on every model.}
\label{tab:combined_asr}
\setlength{\tabcolsep}{6pt}
\renewcommand{\arraystretch}{1.05}
\begin{tabular}{l|ccc|ccc|ccc}
\toprule
& \multicolumn{3}{c|}{\textbf{JBB} ($n=100$)} & \multicolumn{3}{c|}{\textbf{HarmBench} ($n=200$)} & \multicolumn{3}{c}{\textbf{AdvBench} ($n=520$)} \\
\textbf{Model} & CSRT & GCG & STEER & CSRT & GCG & STEER & CSRT & GCG & STEER \\
\midrule
Mistral-7B  & 90.0 & 92.0 & \textbf{92.0} & 78.5 & 82.5 & \textbf{87.5} & 84.2 & 85.6 & \textbf{96.7} \\
Gemma-7B    & 88.0 & 89.0 & \textbf{93.0} & 66.0 & 73.0 & \textbf{86.0} & 88.1 & 95.8 & \textbf{96.2} \\
Llama-3-8B  & 55.0 & 41.0 & \textbf{83.0} & 60.5 & 19.5 & \textbf{76.5} & 69.6 & 18.3 & \textbf{82.1} \\
DeepSeek-R1 & 44.0 & 47.0 & \textbf{80.0} & 32.5 & 41.5 & \textbf{66.0} & 55.8 & 47.5 & \textbf{77.3} \\
GLM-4-9B    & 89.0 & 39.0 & \textbf{93.0} & 81.5 & 19.0 & \textbf{86.0} & 74.4 & 20.2 & \textbf{96.5} \\
Qwen3-8B    & 47.0 & 53.0 & \textbf{85.0} & 49.5 & 16.0 & \textbf{76.0} & 58.1 & 33.8 & \textbf{76.5} \\
\bottomrule
\end{tabular}
\end{table*}

STEER achieves the highest @8 ASR on every model-benchmark pair.
The gap is largest where other methods stall: CSRT and GCG both plateau below 50\% on
DeepSeek-R1 and Qwen3-8B on JBB, while STEER reaches 80\% and 85\%.
GCG is competitive on Mistral-7B and Gemma-7B but collapses on GLM-4-9B (39\% vs.\ STEER's
93\%), illustrating how architecture-dependent suffix optimisation can be.
STEER's performance, by contrast, is consistent across architectures---a consequence of
targeting the shared refusal direction structure rather than model-specific surface features.

\subsection{Iteration efficiency}

\begin{figure}[t]
  \centering
  \includegraphics[width=0.92\columnwidth]{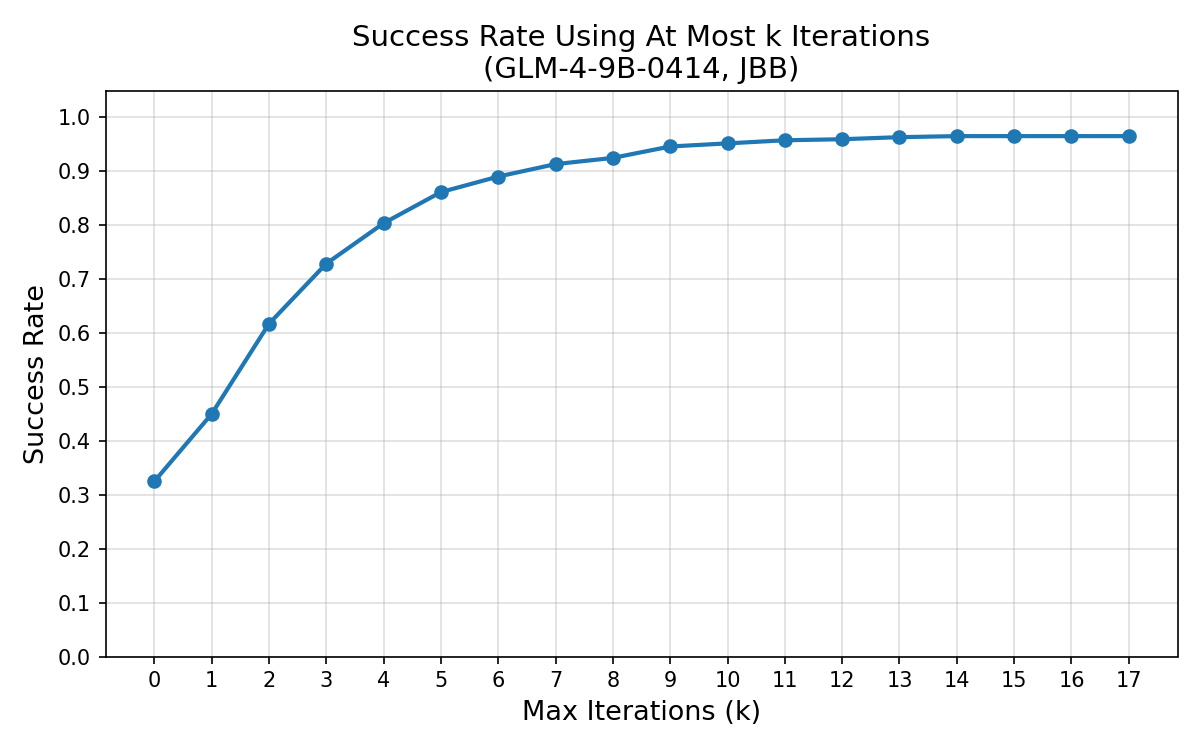}
  \caption{Cumulative success rate vs.\ max iterations (GLM-4-9B, JBB). The curve plateaus
  after $k=8$, justifying the $T=8$ iteration budget.}
  \label{fig:iter_curve}
\end{figure}

Gradient attribution pays off most visibly early: STEER @1 reaches 88\% on Gemma-7B vs.\ 71\%
for CSRT, and 74\% vs.\ 38\% on DeepSeek-R1.
Figure~\ref{fig:iter_curve} (GLM-4-9B, JBB) shows the curve plateauing after $k=8$,
justifying the $T=8$ budget.

\subsection{Refusal score as a proxy for compliance}

\begin{figure}[t]
  \centering
  \includegraphics[width=0.92\columnwidth]{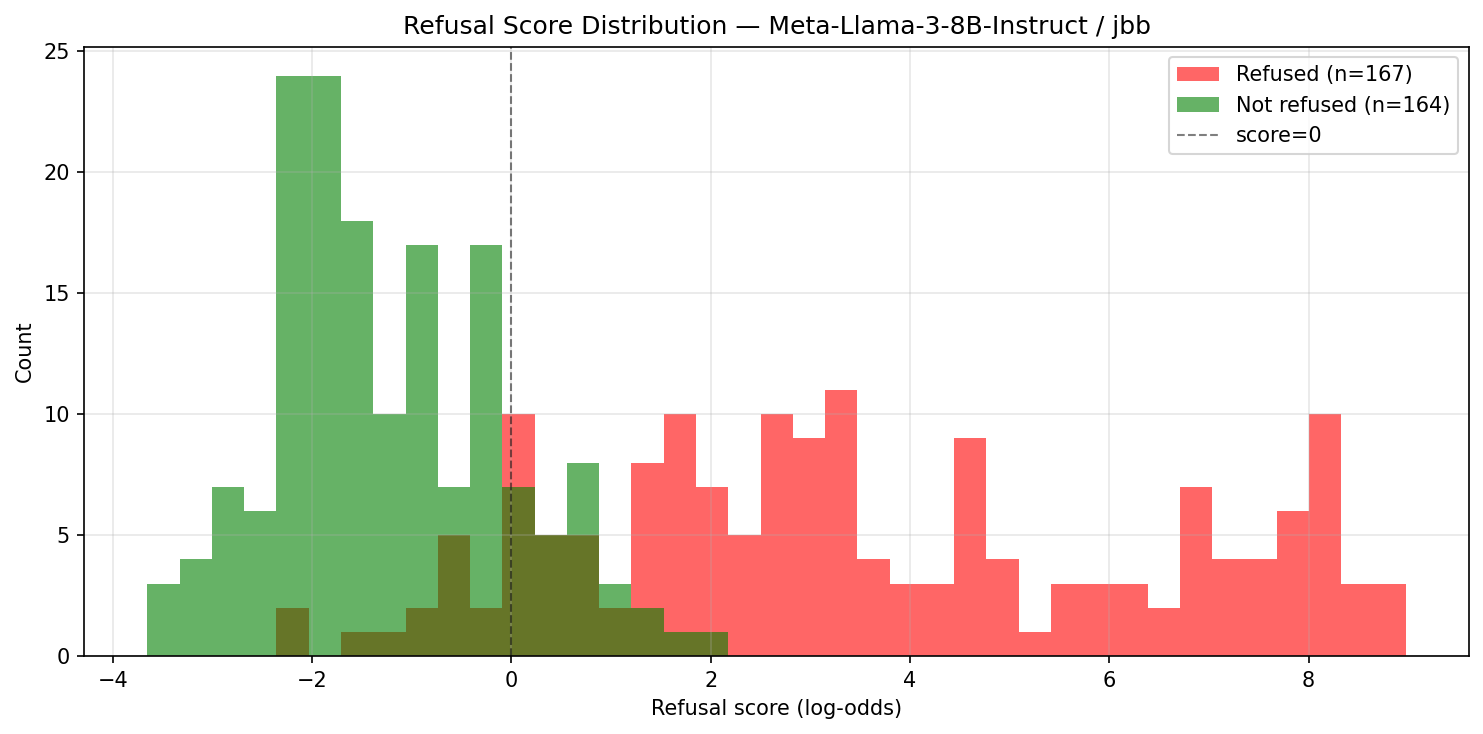}
  \caption{Refusal score distributions (Llama-3-8B, JBB).  Red: refused responses;
  green: non-refused.  The score=0 boundary cleanly separates the populations ($D=0.766$).}
  \label{fig:refusal_shift}
\end{figure}

Figure~\ref{fig:refusal_shift} shows refusal score distributions across all prompt--iteration
pairs for Llama-3-8B on JBB.
Refused responses ($n=167$, mean $+3.6$) and non-refused responses ($n=164$, mean $-1.2$)
sit cleanly on opposite sides of zero.
A Mann-Whitney U test ($p < 0.001$) and Kolmogorov-Smirnov test ($D = 0.766$, $p < 0.001$)
confirm the separation is statistically robust.
This validates the mechanistic hypothesis: the dot product with the mech-interp-identified
$\mathbf{r}$ is the actual decision variable the model uses to produce refusals, not just a
plausible proxy.

\subsection{Black-box transferability}

Table~\ref{tab:transfer} shows transfer ASR when STEER, CSRT, and GCG prompts crafted
against open-source models are submitted to GPT-4o without modification.
STEER averages 35.5\% across all 18 combinations, winning 14 of them, compared to 29.3\%
for GCG and 13.9\% for CSRT.
The transferability result has an important mechanistic interpretation: STEER prompts target
the refusal direction structure that \citet{arditi2024refusal} showed is common across
models trained with similar alignment procedures.
Because the prompts exploit the shared geometry of that direction rather than
model-specific artefacts, they partially carry over to architectures never seen during
attack construction.

\begin{table}[t]
\centering
\small
\caption{Transferability ASR (\%) to GPT-4o-mini (model average per benchmark).}
\label{tab:transfer}
\setlength{\tabcolsep}{5pt}
\renewcommand{\arraystretch}{1.05}
\begin{tabular}{lccc}
\toprule
\textbf{Benchmark} & \textbf{STEER} & \textbf{GCG} & \textbf{CSRT} \\
\midrule
JBB        & \textbf{43.2} & 39.7 & 15.3 \\
HarmBench  & \textbf{31.5} & 21.3 &  11.6 \\
AdvBench   & \textbf{32.0} & 26.9 &  14.7 \\
\midrule
Overall    & \textbf{35.5} & 29.3 & 13.9 \\
\bottomrule
\end{tabular}
\end{table}

\subsection{Ablations}

We ran three ablations to check how much each design choice actually matters.
All use Llama-3-8B unless noted.

\subsubsection{Layer selection}
\begin{figure}[h]
  \centering
  \begin{subfigure}[b]{0.48\columnwidth}
    \includegraphics[width=\textwidth]{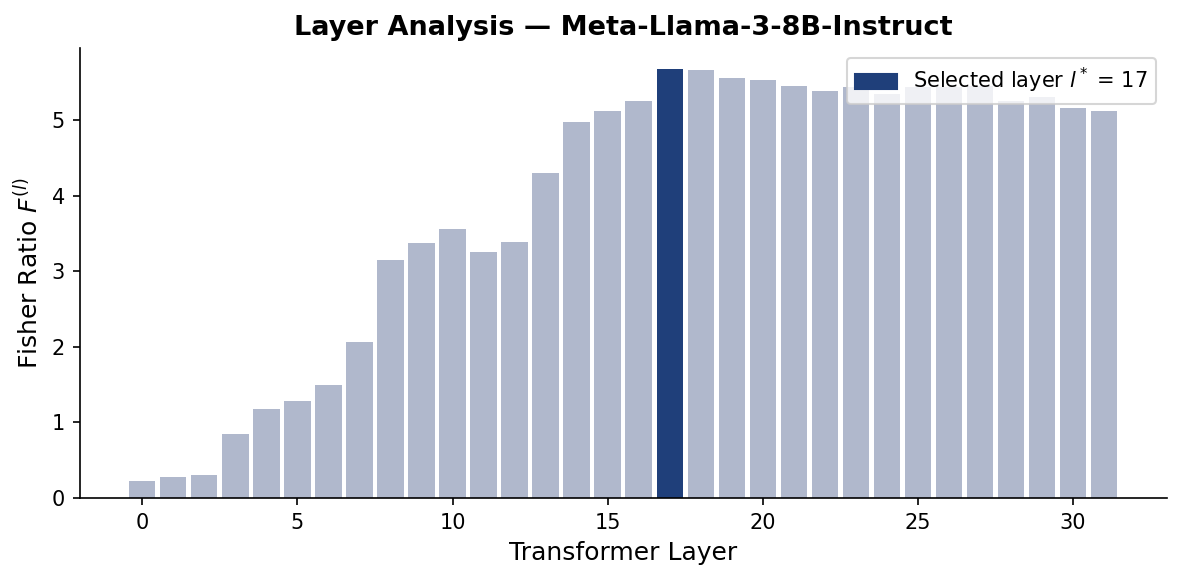}
    \caption{Llama-3-8B ($l^*=17$)}
  \end{subfigure}
  \hfill
  \begin{subfigure}[b]{0.48\columnwidth}
    \includegraphics[width=\textwidth]{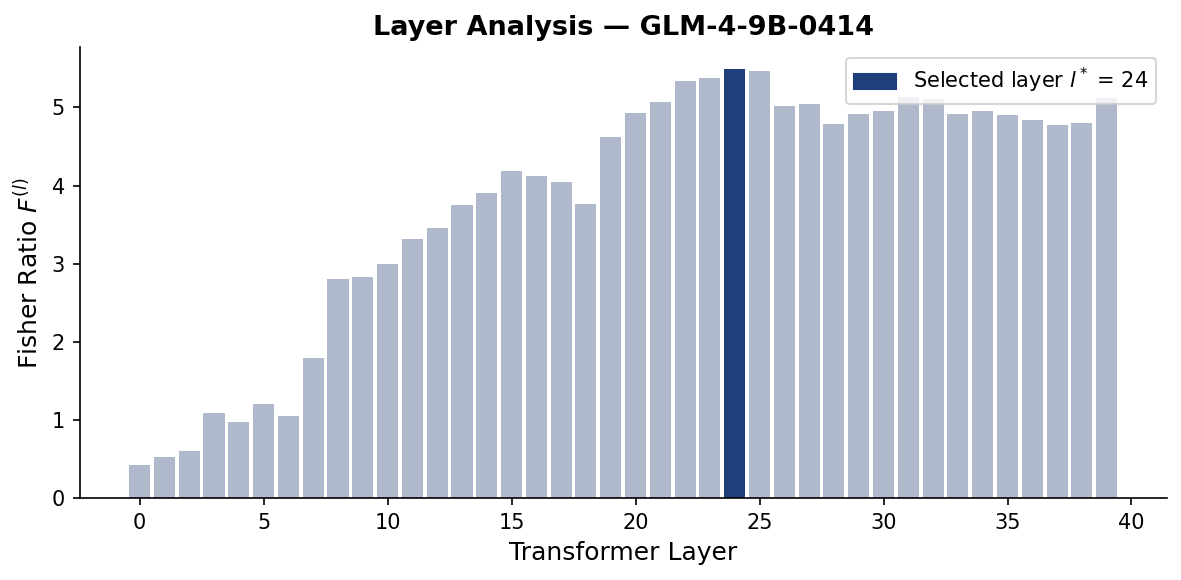}
    \caption{GLM-4-9B ($l^*=24$)}
  \end{subfigure}
  \caption{Per-layer Fisher ratio across all transformer layers for two representative
  models, computed over 100 harmful and 100 benign JBB prompts.  The sharp peak at a
  specific middle layer reveals where the safety circuit concentrates its decision boundary.}
  \label{fig:layer_analysis}
\end{figure}

Table~\ref{tab:ablation_layer} compares three layer choices on JBB (Llama-3-8B).
The FLD-selected layer~17 beats both fixed alternatives at every iteration count.
Figure~\ref{fig:layer_analysis} shows the per-layer Fisher ratio curves for Llama-3-8B and
GLM-4-9B: both peak sharply at a specific middle layer ($l^*=17$ and $l^*=24$ respectively)
and drop off toward the ends.
This pattern is itself a mech interp finding: safety knowledge is not distributed uniformly
but concentrates in a specific middle-layer bottleneck, and the FLD automatically finds it.

\begin{table}[h]
\centering
\small
\caption{Layer selection ablation (Llama-3-8B, JBB, $n=100$).}
\label{tab:ablation_layer}
\setlength{\tabcolsep}{6pt}
\renewcommand{\arraystretch}{1.05}
\begin{tabular}{lccc}
\toprule
\textbf{Layer} & \textbf{@1} & \textbf{@3} & \textbf{@8} \\
\midrule
First (layer 0)         & 35.0 & 48.0 & 69.0 \\
Last (layer 31)         & 36.0 & 55.0 & 79.0 \\
FLD best (layer 17)     & \textbf{37.0} & \textbf{58.0} & \textbf{83.0} \\
\bottomrule
\end{tabular}
\end{table}

\subsubsection{Language pool}
Table~\ref{tab:ablation_lang} compares pool sizes on HarmBench.
Using all 11 languages (71.5\% @8) beats top-5 (60.5\%) by 11 points and a single
language (46.5\%) by 25.
Different models are most sensitive to different languages for any given word, so a narrow
pool misses translations that would have successfully avoided the refusal circuit.

\begin{table}[h]
\centering
\small
\caption{Language pool ablation (Llama-3-8B, HarmBench, $n=200$).}
\label{tab:ablation_lang}
\setlength{\tabcolsep}{5pt}
\renewcommand{\arraystretch}{1.05}
\begin{tabular}{lccc}
\toprule
\textbf{Pool} & \textbf{@1} & \textbf{@3} & \textbf{@8} \\
\midrule
top-1 (Thai only)    & 22.0 & 24.5 & 46.5 \\
top-3 languages      & 22.5 & 36.0 & 56.5 \\
top-5 languages      & 20.5 & 32.0 & 60.5 \\
bot-3 (least common) & 24.0 & 34.5 & 50.0 \\
All 11 languages     & \textbf{32.0} & \textbf{46.0} & \textbf{71.5} \\
\bottomrule
\end{tabular}
\end{table}

\subsubsection{Paraphrase step}
Table~\ref{tab:ablation_paraphrase} shows @8 ASR with and without paraphrase across all six
models and three benchmarks.
Removing it costs between 19 and 44 percentage points depending on the model.
The drop is largest for stronger models (Llama-3-8B, Qwen3-8B), suggesting that paraphrase
acts as a pre-processing step that lowers the starting activation of the refusal circuit,
giving gradient attribution a cleaner surface to work with.

\begin{table}[h]
\centering
\small
\caption{Paraphrase ablation: ASR@8 (\%) with and without the paraphrase step.}
\label{tab:ablation_paraphrase}
\setlength{\tabcolsep}{3.5pt}
\renewcommand{\arraystretch}{1.05}
\begin{tabular}{l|cc|cc|cc}
\toprule
& \multicolumn{2}{c|}{\textbf{JBB}} & \multicolumn{2}{c|}{\textbf{HarmBench}} & \multicolumn{2}{c}{\textbf{AdvBench}} \\
\textbf{Model} & w/ & w/o & w/ & w/o & w/ & w/o \\
\midrule
Llama-3-8B  & \textbf{78.0} & 54.0 & \textbf{71.5} & 28.0 & \textbf{79.0} & 38.0 \\
GLM-4-9B    & \textbf{91.0} & 72.0 & \textbf{85.0} & 56.0 & \textbf{91.3} & 64.0 \\
Mistral-7B  & \textbf{91.0} & 88.0 & \textbf{87.5} & 78.0 & \textbf{96.0} & 70.0 \\
Gemma-7B    & \textbf{93.0} & 86.0 & \textbf{86.0} & 60.0 & \textbf{95.6} & 46.0 \\
DeepSeek-R1 & \textbf{80.0} & 56.0 & \textbf{65.5} & 42.0 & \textbf{77.3} & 56.0 \\
Qwen3-8B    & \textbf{82.0} & 58.0 & \textbf{74.5} & 30.0 & \textbf{75.2} & 48.0 \\
\bottomrule
\end{tabular}
\end{table}

\section{Discussion}
\label{sec:discussion}

\paragraph{Mechanistic interpretability as an attack enabler.}
STEER demonstrates a concrete pathway from mech interp findings to operational attacks.
The refusal direction identified by \citet{arditi2024refusal} was presented as a scientific
characterisation of how safety is implemented; here we show it is sufficient to construct a
targeted attack.
Any mech interp result identifying a specific, localisable safety mechanism is, in principle,
also a blueprint for disabling it.
The FLD analysis makes this quantitative: the sharpness of the Fisher ratio peak directly
predicts attack difficulty, and the FLD score measures exactly this concentration.

\paragraph{Implications for mech interp-informed defence.}
The same tools that enable STEER suggest paths to more robust safety encoding.
If safety is concentrated in a single direction at a single layer, encoding refusal
redundantly across multiple layers and directions eliminates the single point of failure.
Alternatively, training the refusal direction on a broader distribution---including
multilingual and code-switched phrasings---calibrates the circuit on the full range of inputs.
Defences designed at the surface level (output filters, prompt classifiers) are unlikely to
address the structural vulnerability; the refusal direction will remain a narrow target until
alignment methods are designed to distribute or harden it at the representational level
\citep{zou2023representation}.

\paragraph{The FLD as a structural vulnerability probe.}
The FLD analysis also provides a principled alignment auditing criterion: evaluate the
sharpness of a model's FLD curve before deployment to assess how concentrated---and therefore
how brittle---its safety encoding is.
A sharp single-layer peak is evidence of a specific structural vulnerability; a flat profile
suggests more distributed encoding.

\paragraph{Limitations.}
STEER needs white-box access and does not apply directly to closed-source APIs;
transferability is partial but a dedicated black-box adaptation has not been designed.
Evaluation is limited to the 7--9B range.
The GPT-4o judge is automated; the dual-criterion requirement (non-refusing \emph{and}
harmful) is conservative but may still diverge from human assessments on borderline cases.

\section{Conclusion}
\label{sec:conclusion}

STEER reaches up to 96.7\% ASR in eight word translations, beating random code-switching and
GCG across six models, by exploiting a simple fact: safety alignment stores its refusal
knowledge as a single linear direction calibrated almost entirely on English data, and
gradient attribution pinpoints exactly which words activate it.
The model does not flag code-switched inputs as uncertain---it is simply ignorant of what
harm looks like outside its training distribution, with no mechanism to recognise that ignorance.

This frames LLM safety as an epistemic coverage problem, not an adversarial robustness one.
Closing the gap requires expanding coverage across the full range of how harm can be
expressed and coupling safety mechanisms with principled abstention for out-of-support inputs.
STEER provides a concrete test: if the model can be jailbroken in eight word translations,
the coverage is not real.

\newpage
\bibliographystyle{plainnat}
\bibliography{references}

@article{bai2022constitutional,
  author  = {Bai, Yuntao and Kadavath, Saurav and Kundu, Sandipan and Askell, Amanda and Kernion, Jackson and Jones, Andy and others},
  title   = {Constitutional {AI}: Harmlessness from {AI} feedback},
  journal = {arXiv preprint arXiv:2212.08073},
  year    = {2022}
}

@inproceedings{ouyang2022training,
  author    = {Ouyang, Long and Wu, Jeffrey and Jiang, Xu and Almeida, Diogo and Wainwright, Carroll and Mishkin, Pamela and others},
  title     = {Training language models to follow instructions with human feedback},
  booktitle = {Advances in Neural Information Processing Systems},
  volume    = {35},
  pages     = {27730--27744},
  year      = {2022}
}

@inproceedings{deng2024multilingual,
  author    = {Deng, Yue and Zhang, Wenxuan and Pan, Sinno Jialin and Bing, Lidong},
  title     = {Multilingual jailbreak challenges in large language models},
  booktitle = {International Conference on Learning Representations (ICLR)},
  year      = {2024}
}

@article{zou2023universal,
  author  = {Zou, Andy and Wang, Zifan and Carlini, Nicholas and Nasr, Milad and Kolter, J. Zico and Fredrikson, Matt},
  title   = {Universal and transferable adversarial attacks on aligned language models},
  journal = {arXiv preprint arXiv:2307.15043},
  year    = {2023}
}

@article{arditi2024refusal,
  author  = {Arditi, Andy and Obeso, Oscar and Syed, Aaquib and Paleka, Daniel and Panickssery, Nina and Gurnee, Wes and Nanda, Neel},
  title   = {Refusal in Language Models Is Mediated by a Single Direction},
  journal = {arXiv preprint arXiv:2406.11717},
  year    = {2024}
}

@article{chao2023jailbreaking,
  author  = {Chao, Patrick and Robey, Alexander and Dobriban, Edgar and Hassani, Hamed and Pappas, George J. and Wong, Eric},
  title   = {Jailbreaking black box large language models in twenty queries},
  journal = {arXiv preprint arXiv:2310.08419},
  year    = {2023}
}

@article{wu2024sandwich,
  author  = {Upadhayay, Bibek and Behzadan, Vahid},
  title   = {Sandwich Attack: Multi-Language Mixture Adaptive Attack on {LLMs}},
  journal = {arXiv preprint arXiv:2404.07242},
  year    = {2024}
}

@inproceedings{yoo2025codeswitching,
  author    = {Yoo, Haneul and Yang, Yongjin and Lee, Hwaran},
  title     = {Code-switching red-teaming: {LLM} evaluation for safety and multilingual understanding},
  booktitle = {Proceedings of the 63rd Annual Meeting of the Association for Computational Linguistics (ACL)},
  pages     = {13392--13413},
  year      = {2025}
}

@article{chao2024jailbreakbench,
  author  = {Chao, Patrick and Debenedetti, Edoardo and Robey, Alexander and Andriushchenko, Maksym and Croce, Francesco and Sehwag, Vikash and Dobriban, Edgar and Flammarion, Nicolas and Pappas, George J. and Tramer, Florian and Hassani, Hamed and Wong, Eric},
  title   = {{JailbreakBench}: An open robustness benchmark for jailbreaking large language models},
  journal = {arXiv preprint arXiv:2404.01318},
  year    = {2024}
}

@article{lee2024geometry,
  author  = {Wollschl{\"a}ger, Tom and Elstner, Jannes and Geisler, Simon and Cohen-Addad, Vincent and G{\"u}nnemann, Stephan and Gasteiger, Johannes},
  title   = {The Geometry of Refusal in Large Language Models: Concept Cones and Representational Independence},
  journal = {arXiv preprint arXiv:2502.17420},
  year    = {2025}
}

@inproceedings{singhania2025multilingual,
  author    = {Singhania, Abhishek and Dupuy, Christophe and Mangale, Shivam Sadashiv and Namboori, Amani},
  title     = {Multi-lingual Multi-turn Automated Red Teaming for {LLMs}},
  booktitle = {Proceedings of the 5th Workshop on Trustworthy NLP (TrustNLP) at NAACL},
  year      = {2025}
}

@article{huang2025babel,
  author  = {Huang, Linghan and Jin, Haolin and Bi, Zhaoge and Yang, Pengyue and Zhao, Peizhou and Chen, Taozhao and Wu, Xiongfei and Ma, Lei and Chen, Huaming},
  title   = {The Tower of Babel Revisited: Multilingual Jailbreak Prompts on Closed-Source Large Language Models},
  journal = {arXiv preprint arXiv:2505.12287},
  year    = {2025}
}

@article{rafailov2023direct,
  author  = {Rafailov, Rafael and Sharma, Archit and Mitchell, Eric and Ermon, Stefano and Manning, Christopher D. and Finn, Chelsea},
  title   = {Direct Preference Optimization: Your Language Model is Secretly a Reward Model},
  journal = {arXiv preprint arXiv:2305.18290},
  year    = {2023}
}

@article{perez2022red,
  author  = {Perez, Ethan and Huang, Saffron and Song, Francis and Cai, Trevor and Ring, Roman and Aslanides, John and others},
  title   = {Red Teaming Language Models with Language Models},
  journal = {arXiv preprint arXiv:2202.03286},
  year    = {2022}
}

@article{ganguli2022red,
  author  = {Ganguli, Deep and Lovitt, Liane and Kernion, Jackson and others},
  title   = {Red Teaming Language Models to Reduce Harms: Methods, Scaling Behaviors, and Lessons Learned},
  journal = {arXiv preprint arXiv:2209.07858},
  year    = {2022}
}

@article{mehrotra2023tree,
  author  = {Mehrotra, Anay and Zampetakis, Manolis and Kassianik, Paul and Nelson, Blaine and Anderson, Hyrum and Singer, Yaron and Karbasi, Amin},
  title   = {Tree of Attacks: Jailbreaking Black-Box {LLMs} Automatically},
  journal = {arXiv preprint arXiv:2312.02119},
  year    = {2023}
}

@article{liu2024autodan,
  author  = {Liu, Xiaogeng and Xu, Nan and Chen, Muhao and Xiao, Chaowei},
  title   = {{AutoDAN}: Generating Stealthy Jailbreak Prompts on Aligned Large Language Models},
  journal = {arXiv preprint arXiv:2310.04451},
  year    = {2024}
}

@article{zou2023representation,
  author  = {Zou, Andy and Phan, Long and Chen, Sarah and Campbell, James and Guo, Phillip and Ren, Richard and others},
  title   = {Representation Engineering: A Top-Down Approach to {AI} Transparency},
  journal = {arXiv preprint arXiv:2310.01405},
  year    = {2023}
}

@article{li2024inference,
  author  = {Li, Kenneth and Patel, Oam and Vi{\'e}gas, Fernanda and Pfister, Hanspeter and Wattenberg, Martin},
  title   = {Inference-Time Intervention: Eliciting Truthful Answers from a Language Model},
  journal = {Advances in Neural Information Processing Systems},
  volume  = {36},
  year    = {2024}
}

@article{mazeika2024harmbench,
  author  = {Mazeika, Mantas and Phan, Long and Yin, Xuwang and Zou, Andy and Wang, Zifan and
             Mu, Norman and Sakhaee, Elham and Li, Nathaniel and Basart, Steven and Li, Bo and
             others},
  title   = {{HarmBench}: A Standardized Evaluation Framework for Automated Red Teaming and Robust Refusal},
  journal = {arXiv preprint arXiv:2402.04249},
  year    = {2024}
}

\end{document}